\pdfoutput=1

\documentclass[11pt]{article}

\usepackage[review]{ACL2023}

\usepackage{times}
\usepackage{latexsym}

\usepackage{booktabs}

\usepackage{algorithm}
\usepackage{algpseudocode}

\usepackage{tabularx}

\usepackage{multirow}

\usepackage{float}

\usepackage[figuresright]{rotating}

\usepackage{caption}
\usepackage{subcaption}

\usepackage[T1]{fontenc}

\usepackage[utf8]{inputenc}

\usepackage{microtype}

\usepackage{inconsolata}

\title{Toward Unified Controllable Text Generation via Regular Expression Instruction}

\author{Xin Zheng${}^{1,3}$, Hongyu Lin${}^{1}$\thanks{~ Corresponding Authors}, Xianpei Han${}^{1, 2}$\footnotemark[1], Le Sun${}^{1,2}$\\
${}^{1}$Chinese Information Processing Laboratory ~ ${}^{2}$State Key Laboratory of Computer Science \\
Institute of Software, Chinese Academy of Sciences, Beijing, China\\
${}^{3}$University of Chinese Academy of Sciences, Beijing, China \\
{\tt \{zhengxin2020,hongyu,xianpei,sunle\}@iscas.ac.cn} \\
}

\begin{document}
\maketitle
\begin{abstract}

Controllable text generation is a fundamental aspect of natural language generation, with numerous methods proposed for different constraint types. However, these approaches often require significant architectural or decoding modifications, making them challenging to apply to additional constraints or resolve different constraint combinations. To address this, our paper introduces Regular Expression Instruction (REI), which utilizes an instruction-based mechanism to fully exploit regular expressions' advantages to uniformly model diverse constraints. Specifically, our REI supports all popular fine-grained controllable generation constraints, i.e., lexical, positional, and length, as well as their complex combinations, via regular expression-style instructions. Our method only requires fine-tuning on medium-scale language models or few-shot, in-context learning on large language models, and requires no further adjustment when applied to various constraint combinations. Experiments demonstrate that our straightforward approach yields high success rates and adaptability to various constraints while maintaining competitiveness in automatic metrics and outperforming most previous baselines. ~\footnote{Our code and data are available at \url{https://github.com/MrZhengXin/CTG-Regex-Instruction}.}

\end{abstract}

\section{Introduction}

\begin{table}[h!]
\small
\begin{tabular}{p{7cm}}
\midrule
Lexicon \& length constraint \\
\midrule
\textit{Input} \\
\textcolor{red}{<expression>} \textcolor{teal}{<mask\_0>} \textbf{stood}\textcolor{gray}{(0)} \textcolor{teal}{<mask\_1>} \textbf{field}\textcolor{gray}{(1)} \textcolor{teal}{<mask\_2>} \textbf{looking}\textcolor{gray}{(2)} \textcolor{teal}{<mask\_3>} \textcolor{violet}{<length=10>} \textcolor{red}{</expression>} \\
\textit{Output} \\
\textcolor{red}{<expression>} The\textcolor{lightgray}{\_1} player\textcolor{lightgray}{\_2} \textbf{stood}\textcolor{gray}{(0)}\textcolor{lightgray}{\_3} in\textcolor{lightgray}{\_4} the\textcolor{lightgray}{\_5} \textbf{field}\textcolor{gray}{(1)}\textcolor{lightgray}{\_6} \textbf{looking}\textcolor{gray}{(2)}\textcolor{lightgray}{\_7} at\textcolor{lightgray}{\_8} the\textcolor{lightgray}{\_9} batter\textcolor{lightgray}{\_10} \textcolor{red}{</expression>} \\
\midrule

Position \& lexicon constraint \\
\midrule
Input \\
Stephen was at a party. \textcolor{red}{<expression>} \textcolor{teal}{<mask\_0>} \textbf{knocked}\textcolor{gray}{(0)} \textcolor{teal}{<mask\_1>} \textcolor{red}{</expression>} He checked it but it was completely broken. \\
Output \\
\textcolor{red}{<expression>} Stephen \textbf{knocked}\textcolor{gray}{(0)} over a vase while drunk. \textcolor{red}{</expression>} \\
\midrule

Position constraint \& alternative ending  \\
\midrule
\textit{Input} \\
My friends all love to go to the club to dance. They think it's a lot of fun and always invite. I finally decided to tag along last Saturday. \textcolor{red}{<expression>} \textcolor{orange}{<options>} \textcolor{brown}{<choice\_0>} \textcolor{teal}{<mask\_0>} My friends decided to keep inviting me out as I am so much fun. \textcolor{brown}{</choice\_0>} \textcolor{brown}{<choice\_1>} \textcolor{teal}{<mask\_1>} \textbf{The next weekend, I was asked to please stay home.} \textcolor{brown}{</choice\_1>} \textcolor{orange}{</options>} \textcolor{red}{</expression>}
\\
\textit{Output} \\
\textcolor{red}{<expression>} I danced terribly and broke a friend's toe. \textbf{The next weekend, I was asked to please stay home.} \textcolor{red}{</expression>} \\

\midrule
\end{tabular}
\caption{Input and output of instruction prompt based Regular Expression Instruction (REI). REI can describe various types of complex fine-grain constraints, and here we present three examples. Meta-data instruction label is \textcolor{red}{colored}, lexicon constraints or correct choice is \textbf{bold-faced}, and auxiliary marks for length or lexicon uses \textcolor{gray}{gray color}. }

\label{table:expression_example}
\end{table}

Generating texts according to human requirements has long been a critical challenge in natural language generation \citep{DBLP:journals/corr/abs-1909-08593, DBLP:journals/corr/abs-2203-02155}. With the emergence of large language models, many tasks in natural language processing can be unified and converted into the formation of \emph{controllable generation} \citep{prabhumoye-etal-2020-exploring}. For example, text classification \citep{10.1145/183422.183423}, cloze test \citep{devlin-etal-2019-bert}, and multiple-choice question answering \citep{lai-etal-2017-race} tasks constraint the output text to be exactly one of the given options; abductive reasoning \citep{DBLP:conf/iclr/BhagavatulaBMSH20} specifies that the position of the output text is between the previous and future contexts; summarization task \citep{5392697} limits the length of output; machine translation \citep{BARHILLEL196091} demands to use the vocabulary of the target language for text generation.

For controllable text generation, typical fine-grained control tasks include lexicon \citep{lin-etal-2020-commongen}, generating position \citep{shen-etal-2020-blank} and length \citep{carlsson-etal-2022-fine}. Recently, various approaches have been proposed to satisfy these constraints, which can be categorized into three different paradigms: retraining or refactoring the model \citep{DBLP:journals/corr/abs-1909-05858, zhang-etal-2020-pointer, he-2021-parallel, chan2021cocon}; tuning on given data \citep{lester-etal-2021-power, 10.5555/3495724.3495977}; manually designed post-processing \citep{qin-etal-2020-back, qin2022cold, meng2022controllable, lu-etal-2021-neurologic, lu-etal-2022-neurologic, wang-etal-2021-mention}.

Despite the reasonable performance, current methods on transformer-based language models mainly focus on certain constraints but may not be easily transferred to others, let alone the combination of constraints. For example, Non-Residual Prompting \citep{carlsson-etal-2022-fine} and A*esque Decoding \citep{lu-etal-2022-neurologic} only considered lexical and length constraints, but it cannot arbitrarily specify which position the generated text shall occur; on the other hand, COLD \cite{qin2022cold} can generate text given past and future context, but may not add word inclusion constraint nor restrict the output length. Moreover, these controlling methods assume that we have access to the probability distribution or even gradient of the model, but in the case of large language models where we can only obtain the output token via API, these methods may not be available, and thus black-box controlling techniques need further exploration.

To address the above challenges, we proposed instruction-based Regular Expression Instruction (REI), for universal fine-grained controllable generation. Table \ref{table:expression_example} present a few examples. Our instruction design is inspired by regular expression, which can easily describe mainstream constraints and their combinations. Following \citet{rosenbaum-etal-2022-linguist}, we use markup language to construct the expression, hoping that model can better distinguish between meta-data (instructions) and data (actual words). We use two popular paradigms, language model fine-tuning, and large language model few-shot, to teach the model to understand the input constraint expression.

Our method has several advantages. First, our constraint expression supports all typical fine-grained controlling task and is powerful enough to describe composite control specifications. Second, our method can be adapted to various scenarios, such as summarization with length constraint, terminology-constrained machine translation, and alternative-ending story infilling. Third, our method is easy to implement and highly transferrable to other models since it requires only fine-tuning on medium-size models and no further modification on large language models, and it does not need access to probability distribution or gradient.

Experiments demonstrate that current state-of-the-art language models can understand our controlling language, achieving high success rate while maintaining high automatic evaluation metric score and surpassing most of the strong previous baselines under various constraints. We hope our work can shed light on future works. 

\begin{table*}[!htp]
\small

\begin{subtable}{\textwidth}
\centering
\begin{tabular}{p{3.5cm}p{11.5cm}}
\toprule
\textbf{Task} & \textbf{Input with Control Expression} \\
\midrule
$\alpha$NLG & $O_1$ \textcolor{red}{<expression>} \textcolor{teal}{<mask\_0>} \textcolor{red}{</expression>} $O_2$  \\
$\alpha$NLG+length & $O_1$ \textcolor{red}{<expression>} \textcolor{teal}{<mask\_0>} \textcolor{violet}{<length=$l$>} \textcolor{red}{</expression>} $O_2$  \\
$\alpha$NLI & $O_1$ \textcolor{red}{<expression>} \textcolor{orange}{<options>} \textcolor{brown}{<choice\_0>} $H_1$ \textcolor{brown}{</choice\_0>} \textcolor{brown}{<choice\_1>} $H_2$ \textcolor{brown}{</choice\_1>} \textcolor{orange}{</options>} \textcolor{red}{</expression>} $O_2$  \\

CommonGen & \textcolor{red}{<expression>} \textcolor{teal}{<mask\_0>} $c_0$(0) \textcolor{teal}{<mask\_1>} $c_1$(1) \textcolor{teal}{<mask\_2>} $c_2$(2) \textcolor{teal}{<mask\_3>} \textcolor{red}{</expression>}  \\
CommonGen+length & \textcolor{red}{<expression>} \textcolor{teal}{<mask\_0>} $c_0$(0) \textcolor{teal}{<mask\_1>} $c_1$(1) \textcolor{teal}{<mask\_2>} $c_2$(2) \textcolor{teal}{<mask\_3>} \textcolor{violet}{<length=$l$>} \textcolor{red}{</expression>}  \\

\bottomrule
\end{tabular}
\caption{Fine-tune Task}
\label{table:expression_finetune}
\end{subtable}

\begin{subtable}{\textwidth}
\centering
\begin{tabular}{p{3.5cm}p{11.5cm}}
\toprule
\textbf{Task} & \textbf{Input with Control Expression} \\
\midrule

$\alpha$NLG+lexicon & $O_1$ \textcolor{red}{<expression>} \textcolor{teal}{<mask\_0>} $w$(0) \textcolor{teal}{<mask\_1>} \textcolor{red}{</expression>} $O_2$  \\
$\alpha$NLG+length+lexicon & $O_1$ \textcolor{red}{<expression>} \textcolor{teal}{<mask\_0>} $w$(0) \textcolor{teal}{<mask\_1>} \textcolor{violet}{<length=$l$>} \textcolor{red}{</expression>} $O_2$  \\
StoryCompletion+infill & $S_1S_2S_3$ \textcolor{red}{<expression>} \textcolor{teal}{<mask\_0>} \textcolor{orange}{<options>} \textcolor{brown}{<choice\_0>} $E_1$ \textcolor{brown}{</choice\_0>} \textcolor{brown}{<choice\_1>} $E_2$ \textcolor{brown}{</choice\_1>} \textcolor{orange}{</options>} \textcolor{red}{</expression>} \\
Gigaword+length & [Text]\textbackslash n Summarize the aforementioned text in a single phrase.\textbackslash n \textcolor{red}{<expression>} \textcolor{teal}{<mask\_0>} \textcolor{violet}{<length=$l$>} \textcolor{red}{</expression>}  \\
Wiktionary/ITAE & Translate from English to German:\textbackslash n\textbackslash n English: [Text] \textbackslash n German: \textcolor{red}{<expression>} \textcolor{teal}{<mask\_0>} $t_0$(0) \textcolor{teal}{<mask\_1>} $t_1$(1) \textcolor{teal}{<mask\_2>} \textcolor{red}{</expression>} \\

\bottomrule
\end{tabular}
\caption{Transfer Task}
\label{table:expression_transfer}
\end{subtable}

\caption{Constraint expression of each task. We fine-tune on tasks and variations listed in Table \ref{table:expression_finetune}, and additionally evaluate the unseen tasks listed in Table \ref{table:expression_transfer}. Notice that for few-shot learning, all the tasks are not trained before. }
\label{table:expression_by_task}
\end{table*}

\section{Method}
\subsection{Instruction Design}
The controlling language REI follows the style of regular expression due to its expressiveness. Also, it's easy to evaluate whether the input expression instruction matches the generated text or not. Following \citet{rosenbaum-etal-2022-linguist}, HTML-like markup language is used, which helps the model learn that they are meaningful meta-data instructions rather than plain symbols, especially when using large language models in-context learning with limited examples and no parameter update. This markup label can also avoid the usage of the escape character.

REI contains several special labels, as shown in Table \ref{table:expression_example}. \texttt{<expression>} and \texttt{</expression>} mark the beginning and the end of the expression and can be put anywhere in the input text, assuming we only generate according to one expression at a time. \texttt{<mask\_i>} is equivalent to the regular expression ``\texttt{.*}'' and similar to the mask token in BART \citep{lewis-etal-2020-bart} and T5 \citep{10.5555/3455716.3455856}, where at its position the model shall generate zero or more tokens. \texttt{<options>} and \texttt{</options>} is equivalent to the parentheses ``\texttt{(}'' and ``\texttt{)}'' in regular expression, the model shall choose one expression among the group. To make the recognition easier, we use \texttt{<choice\_i>} and  \texttt{</choice\_i>} to wrap each choice. The regular expression notation of length counts at the character level, but in practice, we want to control the output word length. Therefore, we use the \texttt{<length=n>} label to denote the constraint of output word count.

We avoid the shortcoming of T5 \citep{10.5555/3455716.3455856} span-corruption schema, where the model only generates discontinued spans rather than full natural sentences \citep{lester-etal-2021-power}. On the other hand, we also overcome the redundancy of BART denoising schema \citep{he-2021-parallel}, where the whole input is generated again, since we only generate the realized expression. Moreover, beyond fill-in-the-blank, we introduce choice-making, which further enriches the expressiveness of our controlling language.

\subsection{Training}
\label{section:Training}

\paragraph{Fine-tuning} We could automatically construct the training data from the corpus and conduct self-supervised learning. Alternatively, we could also directly convert the input of existing supervised datasets into the form of our controlling language, and use them to fine-tune state-of-the-art models such as FLAN-T5 \citep{DBLP:journals/corr/abs-2210-11416}. The input format is shown in Table \ref{table:expression_finetune}.

We include $\alpha$NLG \citep{DBLP:conf/iclr/BhagavatulaBMSH20} and CommonGen \citep{lin-etal-2020-commongen}, two English controllable generation datasets of position and lexicon constraint. In $\alpha$NLG, given the past observation $O_1$ and the future observation $O_2$, the goal is to generate a hypothesis $h$ that could follow $O_1$ and trigger $O_2$. The regular expression of the constraint is ``\texttt{.*}'' since no lexicon constraint is required. In CommonGen, given a set of $k$ concepts $C=\{c_0, c_1, ..., c_{k-1}\}$, the output text shall include those concepts and at the same time be consistent with common sense. While in the original setting, the appearance order of concepts and their word sense change is not provided, and the model shall make these decisions, here in our controlling language, the exact word and order must be given. Otherwise, we cannot construct the corresponding expression. So, we preprocess the original instances and recover the order and word sense of the concepts by the reference text. To help the model generate the concepts sequentially and track how many concepts it has already used, we append the serial number label $(i)$ to every concept $c_i$ on both the input and output sides and remove the labels from the output generation once completed. The regular expression of the constraint is ``\texttt{.*$c_0$.*$c_1$ ... .*$c_{k-1}$.*}''.

We also leverage these two datasets to teach the model to control the output length by simply adding the length label with the ground truth length. To better track how many words the model itself has already generated, we append the length number label $\_i$ to every word $w_i$; for example, the sentence ``Stephen knocked over a vase while drunk.'' becomes ``Stephen\_0 knocked\_1 over\_2 a\_3 vase\_4 while\_5 drunk.\_6''. Similarly, we remove the length number labels after completion.

Finally, we need to teach the model about choosing grammar. We use $\alpha$NLI \citep{DBLP:conf/iclr/BhagavatulaBMSH20} dataset, the task of which is to determine whether $H_1$ or $H_2$ is the more plausible hypothesis given the past and future observations $O_1$ and $O_2$, and the constraint of the regular expression is ``\texttt{($H_1$|$H_2$)}''.

\paragraph{In-context Learning}
For large language models like GPT-3.5 \citep{NEURIPS2020_1457c0d6}, where typically access is typically provided via API, we may not apply many traditional controllable generation technics. However, we can leverage its ability of in-context learning to conduct fine-grain constraint generation. More specifically, we leverage the ability to discover and imitate the repeated pattern \citep{DBLP:journals/corr/abs-2209-07686, DBLP:journals/corr/abs-2202-12837}, which is desirable in our case, since unlike other natural language understanding tasks, the specific fine-grain constraint is a well-defined simple pattern that could be easily discoverable and imitable.

Given the input with control expression, we can select $k$ instances with the same expression structure as the instruction prompt and send it to the large language model together with input. Naturally, when evaluating the test set, we can select examples from the training set or validation set, or other instances of the test set when they are not available. Consistantly, we use the same input and output format described before, which saves extra efforts on prompt engineering. In addition, we simply use the popular json format ``\texttt{ \{"input": [INPUT], "output": [OUTPUT]\} }'' for each demonstrating instances, and naturally seperate them with ``\textbackslash n''. By using json, we can further avoid the need for escape character if the input text happens to contain metadata like "Input" or "\textbackslash n".

\subsection{Inference}
We use rejection sampling to generate output text that is matched by the control expression. Verifying the output is simple, since we could convert the control expression into regular expression and check the validity. Additionally, if the expression contains length constraint label, we count and compare the number of words in the output text. We try at most $k$ times to avoid infinite loop and save costs if we use large language model API. When using medium or small size langauge model, to increase the generation quality, we can perform beam search first and see if it can generate a valid result at the first try. 

\subsection{Recursive Decoding}
\label{section:recursive_decoding}
Different choice might affect the generated text. For example, consider the case ``\texttt{$S_1$$S_2$$S_3$.*($E_1$|$E_2$)}'', which gives the first three sentence and two alternative endings and the goal is to choose the correct ending while infill the fourth sentence at the same, which is not included in our fine-tuning data. Instead of directly jumping to the answer with possibly insufficient computation, we could also let the model ``think step by step \citep{kojima2022large}''. We can solve each choice expression first, then compare the complete choices ``\texttt{($S_4E_1$|$S_4'E_2$)}"". The generalized decoding procedure is presented at Algorithm \ref{alg:recursive_decoding}, which assumes that each options is independ with each other and greedily solve them from left to right. We leave the evaluation of expression with multipe consecutive options \citep{lu-etal-2022-neurologic} for future work.

\begin{table*}[!htp]
\small
\begin{subtable}{\textwidth}
\centering
\begin{tabular}{p{5cm}llll}
\toprule
\textbf{Method} & \textbf{BLEU} & \textbf{CIDEr} & \textbf{SPICE} & \textbf{Cov.} \\ \midrule
BART \citep{lin-etal-2020-commongen} & 31.83 & 13.96 & 28.00 & 97.35 \\
T5-Large \citep{lin-etal-2020-commongen} & 31.96 & 15.13 & 28.86 & 95.29 \\
Neurologic \citep{lu-etal-2021-neurologic} & 28.10 & 15.50 & 30.80 & 98.50 \\ 
NADO \citep{meng2022controllable} & 30.80 & - & - & 97.10 \\
NRP \citep{carlsson-etal-2022-fine} & - & - & - & 95.10 \\
\midrule
NLI+GPT-3.5, 8 shot, oracle & \textbf{38.89} & \textbf{18.60} & 31.51 & 98.93 \\
REI+GPT-3.5, 8 shot, oracle & 28.64 & 15.15 & 29.49 & 98.60 \\
REI+FLAN-T5-xl, oracle & 36.78 & 18.34 & \textbf{33.56} & \textbf{100.0} \\
\bottomrule
\end{tabular}
\caption{Lexicon constraint}
\label{table:commongen_a}
\end{subtable}

\begin{subtable}{\textwidth}
\centering
\begin{tabular}{p{5cm}llll}
\toprule
\textbf{Method} & \textbf{BLEU} & \textbf{CIDEr} & \textbf{SPICE} & \textbf{SuR.} \\ \midrule
NLI+GPT-3.5, 8 shot  & \textbf{40.48} & \textbf{19.72} & 31.78 & 35.95 \\
REI+GPT-3.5, 8 shot  & 19.53 & 11.54 & 22.35 & 67.43 \\
REI+FLAN-T5-xl & 30.95 & 17.50 & \textbf{32.37} & \textbf{99.90} \\ \bottomrule
\end{tabular}
\caption{Lexicon \& length constraint}
\label{table:commongen_b}
\end{subtable}

\caption{Results on devset of CommonGen. The best models are \textbf{bold} within each metric. }
\label{table:commongen}
\end{table*}

\section{Experiment}

\subsection{Setup}

We conduct experiments on 2 Nvidia A100 GPUs, with about 10 total GPU hours locally. For medium-size language model, we use FLAN-T5-xl \citep{DBLP:journals/corr/abs-2210-11416} with Apache 2.0 license, which has 3B parameters and is fine-tuned on many natural language understanding and generation tasks. We use Huggingface Transformers library \citep{wolf-etal-2020-transformers} with Apache-2.0 license for fine-tuning and evaluation. We trained the model for 3 epochs, with a batch size of 16 and learning rate of 3e-5. We set beam size to 4 for beam search and p to 0.95 for top-p sampling. We generate at most $k=512$ samples if we do not obtain any valid outcome. 

For large language model, we use GPT-3 \citep{NEURIPS2020_1457c0d6} \texttt{text-davinci-003} version via OpenAI API, and the 175B model is calibrated with Reinforcement Learning from Human Feedback \citep{NEURIPS2020_1f89885d}. We feed 8 in-domain examples as the prompt, set the temperature to 0.7, and retry at most $k=8$ times if the result is not valid. All results are from ``single'' run.

\subsection{Lexicon Constraint}
\subsubsection{Lexicon Constraint Only}
\paragraph{Setting} We evaluate our method on the devset of CommonGen \citep{lin-etal-2020-commongen}, as the reference text of the test set is not publicly available. As mentioned in \ref{section:Training}, we feed the model with oracle concept order and word sense. For automatic metrics we use BLEU-4 \citep{papineni-etal-2002-bleu}, CIDEr \citep{Vedantam_2015_CVPR}, SPICE \citep{10.1007/978-3-319-46454-1_24} and Coverage (Cov.), which is the average ratio of input concepts that are present in lemmatizatized outputs.

\paragraph{Results} We compare the performance of our method with other baselines, including fine-tuning methods BART \citep{lin-etal-2020-commongen} and T5-Large \citep{lin-etal-2020-commongen}, auxilary guiding model method NADO \citep{meng2022controllable}, prompting method NRP \citep{carlsson-etal-2022-fine}, and 8-shot pure natural language instruction (NLI) on GPT-3.5, which is shown at Table \ref{table:commongen_a}. 

Given only 8 examples with a clear connection between input and output, GPT-3.5 still shows competitive performance in terms of text automatic metrics, and achieves high concept coverage, surpassing all the previous baselines. Compared with natural language 
instruction, the success rate is very close. And with more supervised data to modify the model's parameter, FLAN-T5-xl performs significantly better than GPT-3.5 and other previous baselines in all metrics and successfully satisfies all lexicon constraints.

\subsubsection{Lexicon \& Length Constraint}
\label{section:lexicon_length_constraint}
As described in Section \ref{section:Training}, we slightly modify the devset of CommonGen to introduce the additional length constraint and evaluate GPT-3.5 and FLAN-T5. For metric, we replace Coverage (Cov.) with Success Rate (SuR.), which is the average percentage of output that matches the input expression. In a composite task, the performance of GPT-3.5 downgrades dramatically and struggles to generate valid output, indicating that multi-concept inclusion and length control at the same time is challenging, especially for few-shot in-context learning. Yet, REI still outperforms NLI in terms of success rate, and the "high" n-gram metrics might also indicate the poor instruction following ability in terms of challenging fine-grain constraints, which is consistent with the finding of \citet{DBLP:journals/corr/abs-2304-14293}. FLAN-T5 only has a minor drop in performance and still maintains a high success rate since it has trained on this composite constraint.

\subsection{Position Constraint}
\subsubsection{Position constraint only}
\paragraph{Setting} We evaluate our method on the testset of $\alpha$NLG \citep{DBLP:conf/iclr/BhagavatulaBMSH20}. The automatic metrics include BLEU-4 \citep{papineni-etal-2002-bleu}, ROUGE-L \citep{lin-2004-rouge} and BERTScore \citep{Zhang*2020BERTScore:}. We do not report Success Rate since it's always 100\%. 

\paragraph{Results}
As presented in Table \ref{table:anlg_a}, we compare our method with two unsupervised baselines DeLorean \citep{qin-etal-2020-back} and COLD \citep{qin2022cold}, non-autoregressive Diffusion-LM \citep{li2022diffusionlm} and two fine-tuned methods on 11B T5 \citep{khashabi2021genie}, 20B UL2 \citep{DBLP:journals/corr/abs-2205-05131} and 8-shot NLI on GPT-3.5.

With few-shot learning, GPT-3.5 outperforms two unsupervised baselines and Diffusion-LM, demonstrating its strong in-context learning ability given only a few infilling examples. Since it's a relatively simple constraint, the performance between REI and NLI is very close. With our careful instruction prompt design and adequate fine-tuning, 3B FLAN-T5 shows stronger performance than 11B T5, and remains competitive compared to 20B UL2.

\begin{table}[h]
\small

\begin{subtable}{0.48\textwidth}
\centering
\begin{tabular}{p{2.91cm}llp{1cm}}
\toprule

\textbf{Method} & \textbf{BLEU} & \textbf{ROUGE} & \textbf{BERT} \\
\midrule

\citet{qin-etal-2020-back} & 1.38 & 18.94 & 42.86 \\

\citet{qin2022cold} & 1.79 & 19.50 & 42.67 \\

\citet{li2022diffusionlm} & 7.10 & 28.30 & 89.00 \\

\citet{khashabi2021genie} & 19.47 & 44.60 & 92.87 \\

\citet{DBLP:journals/corr/abs-2205-05131} & 24.34 & \textbf{49.30} & \textbf{93.51} \\

\midrule
NLI+GPT-3.5, 8 shot  & 13.62 & 36.38 & 91.05  \\
REI+GPT-3.5, 8 shot  & 13.01 & 37.29 & 91.27 \\
REI+FLAN-T5-xl & \textbf{25.44} & 48.45 & 93.28 \\
\bottomrule
\end{tabular}
\caption{Position constraint}
\label{table:anlg_a}
\end{subtable}

\begin{subtable}{0.48\textwidth}
\centering
\begin{tabular}{p{2.91cm}llp{1cm}}
\toprule
\textbf{Model} & \textbf{BLEU} & \textbf{ROUGE} & \textbf{SuR.} \\
\midrule
NLI+GPT-3.5, 8 shot  & 9.9 & 32.93 & 42.09  \\
REI+GPT-3.5, 8 shot  & 10.63 & 34.87 & 96.80 \\
REI+FLAN-T5-xl & \textbf{19.92} & \textbf{46.17} & \textbf{100.0} \\
\bottomrule
\end{tabular}
\caption{Position \& length constraint}
\label{table:anlg_b}
\end{subtable}

\begin{subtable}{0.48\textwidth}
\centering
\begin{tabular}{p{2.91cm}llp{1cm}}
\toprule
\textbf{Model} & \textbf{BLEU} & \textbf{ROUGE} & \textbf{SuR.} \\
\midrule
NLI+GPT-3.5, 8 shot  & 14.76 & 42.04 & 99.01  \\
REI+GPT-3.5, 8 shot  & 18.59 & 44.67 & 99.44 \\
REI+FLAN-T5-xl & \textbf{23.56} & \textbf{48.81} & \textbf{99.78} \\
\bottomrule
\end{tabular}
\caption{Position \& lexicon constraint}
\label{table:anlg_c}
\end{subtable}

\begin{subtable}{0.48\textwidth}
\centering
\begin{tabular}{p{2.91cm}llp{0.9cm}}
\toprule
\textbf{Model} & \textbf{BLEU} & \textbf{ROUGE} & \textbf{SuR.} \\
\midrule
NLI+GPT-3.5, 8 shot  & 19.14 & 43.67 & 28.00  \\
REI+GPT-3.5, 8 shot  & 17.45  & 43.90  & 94.02 \\
REI+FLAN-T5-xl & \textbf{21.99} & \textbf{49.17} & \textbf{99.69} \\
\bottomrule
\end{tabular}
\caption{Position \& length \& lexicon constraint}
\label{table:anlg_d}
\end{subtable}

\caption{Result on test of $\alpha$NLG.}
\label{table:anlg}
\end{table}

\subsubsection{Position \& Length Constraint}
As mentioned in Section \ref{section:Training}, we slightly modify the $\alpha$NLG test set to add the length constraint. We change the BERTScore metric to SuccessRate (SuR.). Table \ref{table:anlg_b} shows the results. GPT-3.5 manages to imitate both position and length constraints, showing relatively high success rate, while under NLI, it performs badly. But with full-scale supervised learning, FLAN-T5 can robustly generate valid output on the test set 100\% of the time. Also, in terms of automatic metrics, the output of both models does not downgrade dramatically. 
\subsubsection{Position \& Lexicon Constraint}
We can also modify the $\alpha$NLG test set to add lexicon constraint, setting the keyword to be the first verb on the reference text. The input format is shown in Table \ref{table:expression_transfer}, and Table \ref{table:anlg_c} shows the results. For GPT-3.5, it still is very likely to generate valid output nearly all of the time, and the automatic metrics enjoy improvement compared with the results of no lexicon constraint, since the additional gold words are provided, and the verb constraint limits the vast scope of possible hypothesis space. Also, REI is slightly better than NLI. For FLAN-T5, although it has been trained on position constraint or lexicon constraint separately, it has not seen the combination, and yet still demonstrates strong performance.
\subsubsection{Position \& Lexicon \& Length Constraint}
\label{section:position_lexicon_length_constraint}
We can further combine all conditions together, adding both length and lexicon constraints on the test set of $\alpha$NLG. The input format is presented in Table \ref{table:expression_transfer}, and Table \ref{table:anlg_d} shows the results. Compositional constraints challenge few-shot GPT-3.5, as it's more difficult to generate output that matches all three requirements, and the success rate drops slightly. Interestingly, NLI got a very low success rate. But fully-trained FLAN-T5 exhibits robust transfer ability, as the simultaneous three constraints are not included in training data, but FLAN-T5 still manages to achieve close to 100\% success rate. 

\subsubsection{Position Constraint \& Alternative Endings}
\label{section:position_alternative}
On the test set of Story Cloze Test \citep{mostafazadeh-etal-2016-corpus}, which is to choose between the right ending and the wrong one given the four-sentence context, we additionally mask the fourth sentence and require the model to infill the missing sentence while determining the correct ending. The input format is shown in Table \ref{table:expression_transfer}, and the result is shown in Table \ref{table:story_completion_infill}. We change the Success Rate (SuR.) metric to Accuracy (Acc.), since choosing either ending is valid. For GPT-3.5, we directly construct promoting examples with the initial input and final output, and surprisingly find that GPT-3.5 handles the composite constraint quite well, and chooses the right ending with not bad accuracy. Also, REI comes close to NLI in performance. For FLAN-T5-xl, we use the recursive decoding (Section \ref{section:recursive_decoding}, and it shows moderate performance, with lower accuracy but higher BLEU / ROUGE compared with GPT-3.5.

\begin{table*}[!tp]
\small
\centering
\begin{tabular}{lllll}
\toprule
 & \multicolumn{2}{c}{\textbf{Wiktionary}} & \multicolumn{2}{c}{\textbf{IATE}} \\
\multirow{-2}{*}{\textbf{Method}} & \textbf{Term\%} & \textbf{BLEU} & \textbf{Term\%} & \textbf{BLEU} \\
\midrule

Constraint decoding \citep{dinu-etal-2019-training} & 99.50 & 25.80 & 82.00 & 25.30 \\
Train-by-replace \citep{dinu-etal-2019-training} & 93.40 & 26.30 & 94.50 & 26.00 \\
RePP \citep{DBLP:journals/corr/abs-2209-11409} & 93.67 & 30.52 & 95.41 & 29.38 \\

TADA \citep{ailem-etal-2021-encouraging} & 96.84 & 26.73 & 98.02 & 27.11 \\
EDITOR \citep{10.1162/tacl_a_00368} & 99.8 & 29.30 & \textbf{100.0} & 28.90 \\
Levenshtein Transformer \citep{susanto-etal-2020-lexically}  & \textbf{100.0} & 31.20 & \textbf{100.0} & 30.13 \\
\midrule
NLI+GPT-3.5, 8-shot & 99.03 & \textbf{37.62} & 98.07 & 32.22 \\
REI+GPT-3.5, 8-shot & 99.52 & 34.88 & 99.45 & \textbf{35.25} \\
\bottomrule
\end{tabular}
\caption{Results on Wiktionary and IATE.}
\label{table:terminology_machine_translation}
\end{table*}

\begin{table}[h]
\small
\centering

\begin{tabularx}{0.45\textwidth}{llll}
\toprule
\textbf{Method} & \textbf{BLEU} & \textbf{ROUGE} & \textbf{Acc.} \\
\midrule
NLI+GPT-3.5, 8 shot  & 3.83 & \textbf{21.27} & \textbf{88.99} \\
REI+GPT-3.5, 8 shot  & 3.77  & 20.56  & 88.72 \\
REI+FLAN-T5-xl & \textbf{3.87}  & 20.9  & 84.61 \\
\bottomrule

\end{tabularx}

\caption{Results on Story Cloze Test with positional constraint.}
\label{table:story_completion_infill}
\end{table}

\subsection{Summarization with length constraint}
\label{section:summarization_length}
\begin{table}[h]
\small
\centering

\begin{tabularx}{0.45\textwidth}{lll}
\toprule
\textbf{Method} & \textbf{ROUGE} & \textbf{SuR.} \\
\midrule
SEQ \citep{baziotis-etal-2019-seq} & 22.68 & - \\
TED \citep{yang-etal-2020-ted} & 22.83 & - \\
\midrule
NLI+GPT-3.5, 8 shot  & 24.62 & 28.87  \\
REI+GPT-3.5, 8 shot  & 25.46 & 79.51 \\
REI+FLAN-T5-xl & \textbf{28.49} & \textbf{100.0} \\
\bottomrule

\end{tabularx}
\caption{Results on the test set of Gigaword.}
\label{table:summarization}
\end{table}
REI can also easily support abstractive summarization with desired length \cite{kikuchi-etal-2016-controlling, fan-etal-2018-controllable}, as long as the base model has been trained on the summarization task, which is the case in our choosing models FLAN-T5 \citep{DBLP:journals/corr/abs-2210-11416} and GPT-3.5 \citep{DBLP:journals/corr/abs-2203-02155}. We choose to evaluate on the test set of English headline generation dataset Gigaword \citep{graff2003english}, due to its short input and output length. Also, Gigaword is not included in the training set of FLAN-T5 or GPT-3.5. The input format is written in Table \ref{table:expression_transfer}. We use ROUGE-L \citep{lin-2004-rouge} and Success Rate (SuR.) for metrics. 

We compare our methods with two unsupervised unconstrainted baselines SEQ \citep{baziotis-etal-2019-seq} and TED \citep{yang-etal-2020-ted}, and the results are shown in Table \ref{table:summarization}. Both GPT-3.5 and FLAN-T5 exceed the two baselines in ROUGE-L score, showing relatively good text quality. Since the summarization task constrains more on the semantic of output compared with pure lexicon constraint (CommonGen) or position constraint ($\alpha$NLG), satisfying length constraint might be more difficult, and GPT-3.5 shows a relatively lower success rate, but NLI has the worst success rate. But nevertheless, FLAN-T5 still achieves 100\% success rate. Notice that with limited REI training tasks, the model can still generalize to new tasks with the specific format, demonstrating the robust transfer ability under supervised learning.

\subsection{Terminology-constrainted machine transaltion}
\label{section:terminology_translation}
We can also apply REI to machine translation with terminology constraint \citep{dinu-etal-2019-training}, which is to ensure the given terminologies $T=(t_0, t_1, ...)$ are used in translation. We only test GPT-3.5 here, due to its superiority in multi-language understanding, while the majority of output language during pre-training, multi-task learning, and fine-tuning is English. We evaluate on the test set of Wiktionary and IATE \citep{dinu-etal-2019-training}, two English-German translation dataset, using BLEU-4 \citep{papineni-etal-2002-bleu} and Terminology Coverage (Term) for metrics.

We compare our method with several strong baselines, including Constraint decoding \citep{dinu-etal-2019-training}, Train-by-replace \citep{dinu-etal-2019-training}, RePP \citep{DBLP:journals/corr/abs-2209-11409}, TADA \citep{ailem-etal-2021-encouraging}, EDITOR \citep{10.1162/tacl_a_00368}, Levenshtein Transformer \citep{susanto-etal-2020-lexically}, and 8-shot NLI on GPT-3.5. Due to its vast parameters, GPT-3.5 outperforms all other baselines in terms of BLEU score. Also, GPT-3.5 achieves near 100\% terminology coverage rate, which is close to the existing upper limit. Finally, REI has a slightly higher term coverage than NLI.

\begin{table*}[!htp]

\small
\centering
\begin{tabular}{p{3cm}p{12cm}}
\hline

CommonGen+length & \textcolor{red}{<expression>} \textcolor{teal}{<mask\_0>} \textbf{dance}\textcolor{gray}{(0)} \textcolor{teal}{<mask\_1>} \textbf{performed}\textcolor{gray}{(1)} \textcolor{teal}{<mask\_2>} \textbf{stage}\textcolor{gray}{(2)} \textcolor{teal}{<mask\_3>} \textbf{wearing}\textcolor{gray}{(3)} \textcolor{teal}{<mask\_4>} \textbf{costumes}\textcolor{gray}{(4)} \textcolor{teal}{<mask\_5>} \textcolor{violet}{<length=11>} \textcolor{red}{</expression>}  \\
FLAN-T5-xl & A\textcolor{lightgray}{\_1} \textbf{dance}\textcolor{gray}{(0)}\textcolor{lightgray}{\_2} is\textcolor{lightgray}{\_3} \textbf{performed}\textcolor{gray}{(1)}\textcolor{lightgray}{\_4} on\textcolor{lightgray}{\_5} a\textcolor{lightgray}{\_6} \textbf{stage}\textcolor{gray}{(2)}\textcolor{lightgray}{\_7} by\textcolor{lightgray}{\_8} people\textcolor{lightgray}{\_9} \textbf{wearing}\textcolor{gray}{(3)}\textcolor{lightgray}{\_10} \textbf{costumes}\textcolor{gray}{(4)}\textcolor{lightgray}{\_11} \\
GPT-3.5, 8 shot & A\textcolor{lightgray}{\_1} traditional\textcolor{lightgray}{\_2} \textbf{dance}\textcolor{gray}{(0)}\textcolor{lightgray}{\_3} is\textcolor{lightgray}{\_4} \textbf{performed}\textcolor{gray}{(1)}\textcolor{lightgray}{\_5} on\textcolor{lightgray}{\_6} the\textcolor{lightgray}{\_7} \textbf{stage}\textcolor{gray}{(2)},\textcolor{lightgray}{\_8} \textbf{wearing}\textcolor{gray}{(3)}\textcolor{lightgray}{\_9} colorful\textcolor{lightgray}{\_10} \textbf{costumes}\textcolor{gray}{(4)}\textcolor{lightgray}{\_11} \\
\hline

$\alpha$NLG+length+lexicon & Jim was not confident in his home repair skills. \textcolor{red}{<expression>} \textcolor{teal}{<mask\_0>} \textbf{attended}\textcolor{gray}{(0)} \textcolor{teal}{<mask\_1>} \textcolor{violet}{<length=9>} \textcolor{red}{</expression>} Jim was so excited to learn a new skill.  \\
FLAN-T5-xl & Jim\textcolor{lightgray}{\_1} bought\textcolor{lightgray}{\_2} new\textcolor{lightgray}{\_3} gloves\textcolor{lightgray}{\_4} and\textcolor{lightgray}{\_5} \textbf{attended}\textcolor{gray}{(0)}\textcolor{lightgray}{\_6} a\textcolor{lightgray}{\_7} home\textcolor{lightgray}{\_8} repair.\textcolor{lightgray}{\_9} \\
GPT-3.5, 8 shot & Jim\textcolor{lightgray}{\_1} \textbf{attended}\textcolor{gray}{(0)}\textcolor{lightgray}{\_2} a\textcolor{lightgray}{\_3} home\textcolor{lightgray}{\_4} repair\textcolor{lightgray}{\_5} workshop\textcolor{lightgray}{\_6} to\textcolor{lightgray}{\_7} gain\textcolor{lightgray}{\_8} confidence.\textcolor{lightgray}{\_9} \\
\hline

StoryCompletion+infill & I tried going to the park the other day. The weather seemed nice enough for a walk. Within minutes of getting there I started sneezing. \textcolor{red}{<expression>} \textcolor{orange}{<options>} \textcolor{brown}{<choice\_0>} \textcolor{teal}{<mask\_0>} \textbf{My allergies were too bad and I had to go back home.} \textcolor{brown}{</choice\_0>} \textcolor{brown}{<choice\_1>} \textcolor{teal}{<mask\_1>} It reminded me of how much I loved spring flowers. \textcolor{brown}{</choice\_1>} \textcolor{orange}{</options>} \textcolor{red}{</expression>} \\
FLAN-T5-xl & There were a lot of people at the park. \textbf{My allergies were too bad and I had to go back home.} \\
GPT-3.5, 8 shot & I realized I had forgotten the antihistamines at home. \textbf{My allergies were too bad and I had to go back home.} \\
\hline
Gigaword+length & japan 's toyota team europe were banned from the world rally championship for one year here on friday in a crushing ruling by the world council of the international automobile federation.\textbackslash n Summarize the aforementioned text in a single phrase.\textbackslash n \textcolor{red}{<expression>} \textcolor{teal}{<mask\_0>} \textcolor{violet}{<length=6>} \textcolor{red}{</expression>}  \\
FLAN-T5-xl & toyota\textcolor{lightgray}{\_1} team\textcolor{lightgray}{\_2} europe\textcolor{lightgray}{\_3} banned\textcolor{lightgray}{\_4} from\textcolor{lightgray}{\_5} rallying\textcolor{lightgray}{\_6} \\
GPT-3.5, 8 shot & toyota\textcolor{lightgray}{\_1} team\textcolor{lightgray}{\_2} europe\textcolor{lightgray}{\_3} banned\textcolor{lightgray}{\_4} by\textcolor{lightgray}{\_5} fia\textcolor{lightgray}{\_6} \\
\midrule
Wiktionary & Translate from English to German:\textbackslash n\textbackslash n English: Jennifer Aniston need not always be perfect or successful. \textbackslash n German: \textcolor{red}{<expression>} \textcolor{teal}{<mask\_0>} \textbf{erfolgreich}\textcolor{gray}{(0)} \textcolor{teal}{<mask\_1>} \textcolor{red}{</expression>} \\
GPT-3.5, 8 shot & Jennifer Aniston muss nicht immer perfekt oder \textbf{erfolgreich}\textcolor{gray}{(0)} sein. \\
\midrule
\end{tabular}

\caption{Qualitative examples of various constraints by fine-tuned FLAN-T5-xl and few-shot GPT-3.5. }
\label{table:qualitative_results}
\end{table*}

\subsection{Qualitative Results}
Table \ref{table:qualitative_results} shows the samples of lexicon \& length constraints (Section \ref{section:lexicon_length_constraint}), position \& lexicon \& length constraints (Section \ref{section:position_lexicon_length_constraint}), position constraint with alternative ending (Section \ref{section:position_alternative}), summarization with length constraint (Section \ref{section:summarization_length}) and translation with terminology constraint (Section \ref{section:terminology_translation}). Both FLAN-T5 and GPT-3.5 generate valid and fluent sentences. GPT-3.5 also uses more vivid or human-like words like ``antihistamines'' or the abbreviation ``FIA'', probably due to its large-scale model size and training corpus.

\section{Related Work}
\paragraph{Tasks of Controllable Text Generation}
Controllable text generation refers to the tasks that generate text according to the controlling signals \citep{prabhumoye-etal-2020-exploring}. Typically, the output can be constrained at three levels from coarse to fine: \citep{DBLP:journals/corr/abs-2201-05337} semantic, structural and lexical. At semantic level, the signals include topic \citep{tang-etal-2019-topic}, sentiment \citep{10.5555/3327345.3327417}, format \citep{li-etal-2020-rigid}, toxity \citep{krause-etal-2021-gedi-generative} and other abstract attribute. At the structural level, the constraints include key-value data table \citep{novikova-etal-2017-e2e}, syntax tree, and parts-of-speech \citep{li2022diffusionlm}. At lexical level, then controlling elements include keyword \citep{lin-etal-2020-commongen}, generating position \citep{shen-etal-2020-blank} and length \citep{carlsson-etal-2022-fine}. 
\paragraph{Methods of Controllable Text Generation}
Current approach for controllable text generation can be summarized as three main categories \citep{DBLP:journals/corr/abs-2201-05337}: retraining or refactoring the model, e.g. CTRL \citep{DBLP:journals/corr/abs-1909-05858}, POINTER \citep{zhang-etal-2020-pointer}, CMDP \citep{chan-etal-2021-controllable}, Constrained BART \citep{he-2021-parallel}, CoCon \citep{chan2021cocon}, PlanGen \citep{su-etal-2021-plan-generate} and InstructCTG \citep{DBLP:journals/corr/abs-2304-14293}; tuning on given data, including model fine-tuning, Prompt Tuning \citep{lester-etal-2021-power} and RL-Fine Tuning \citep{10.5555/3495724.3495977}; and post-processing, which can either design specific decoding strategy, e.g. Constrainted Beam Search \citep{anderson-etal-2017-guided}, DeLorean \citep{qin-etal-2020-back}, COLD \cite{qin2022cold}, NeuroLogic \citep{lu-etal-2021-neurologic}; or using auxilary guiding model, e.g. PPLM \citep{anderson-etal-2017-guided}, GeDI \citep{krause-etal-2021-gedi-generative}, FUDGE \citep{yang-klein-2021-fudge}, CTRLsum \citep{he-etal-2022-ctrlsum}, Plug-and-Play Content Planning \citep{liu-etal-2022-plug}, NADO \citep{meng2022controllable}, and MACSum \citep{10.1162/tacl_a_00575} .

\section{Conclusion}
We proposed Regular Expression Instruction (REI), a novel instruction-based method that unifies fine-grain lexical-level constrained text generation. Our method is highly adaptable, fitting either language model fine-tuning or large language model in-context learning. Our controlling language can also easily be applied to other related tasks, including story completion while infilling, summarization with length constraint, and machine translation with terminology constraint. Experiments show that our method has a high success rate and outperforms most of the previous strong baselines, demonstrating its effectiveness despite the simplicity. We leave the evaluation and improvement of more complex constraints for future works.

\section*{Limitations}
Our proposed Regular Expression Instruction is serialized and cannot describe a set of keyword constraints where the appearing order is arbitrary, but only a list of keywords with determined order. Future work is needed to exceed the limit, either by approximating the word order or by repeated random sampling. Also, to obtain valid results we use reject sampling, which might need many repeated trials, thus reducing the efficiency and downgrading the speed. More efficient mechanisms with less retry are worth investigating. Additionally, under the current trends of the instruction following, more sophisticated prompts under 0-shot is worth investigating.  

\section*{Ethics Statement}
This work involves no sensitive data and uses several public-available datasets. This work discusses controllable text generation, which aims for better usage of the black-box language model and may better reduce the problematic biases. We notice that the method proposed in this work can be used to generate disinformation or harmful content directly via controlling language, but the malicious usage can be further avoided by filtering out improper control input and stopping harmful content generation.

\bibliography{anthology,custom}

\begin{thebibliography}{67}
\expandafter\ifx\csname natexlab\endcsname\relax\def\natexlab#1{#1}\fi

\bibitem[{Ailem et~al.(2021)Ailem, Liu, and
  Qader}]{ailem-etal-2021-encouraging}
Melissa Ailem, Jingshu Liu, and Raheel Qader. 2021.
\newblock \href {https://doi.org/10.18653/v1/2021.findings-acl.125}
  {Encouraging neural machine translation to satisfy terminology constraints}.
\newblock In \emph{Findings of the Association for Computational Linguistics:
  ACL-IJCNLP 2021}, pages 1450--1455, Online. Association for Computational
  Linguistics.

\bibitem[{Anderson et~al.(2016)Anderson, Fernando, Johnson, and
  Gould}]{10.1007/978-3-319-46454-1_24}
Peter Anderson, Basura Fernando, Mark Johnson, and Stephen Gould. 2016.
\newblock Spice: Semantic propositional image caption evaluation.
\newblock In \emph{Computer Vision -- ECCV 2016}, pages 382--398, Cham.
  Springer International Publishing.

\bibitem[{Anderson et~al.(2017)Anderson, Fernando, Johnson, and
  Gould}]{anderson-etal-2017-guided}
Peter Anderson, Basura Fernando, Mark Johnson, and Stephen Gould. 2017.
\newblock \href {https://doi.org/10.18653/v1/D17-1098} {Guided open vocabulary
  image captioning with constrained beam search}.
\newblock In \emph{Proceedings of the 2017 Conference on Empirical Methods in
  Natural Language Processing}, pages 936--945, Copenhagen, Denmark.
  Association for Computational Linguistics.

\bibitem[{Apt\'{e} et~al.(1994)Apt\'{e}, Damerau, and
  Weiss}]{10.1145/183422.183423}
Chidanand Apt\'{e}, Fred Damerau, and Sholom~M. Weiss. 1994.
\newblock \href {https://doi.org/10.1145/183422.183423} {Automated learning of
  decision rules for text categorization}.
\newblock \emph{ACM Trans. Inf. Syst.}, 12(3):233–251.

\bibitem[{Bar-Hillel(1960)}]{BARHILLEL196091}
Yehoshua Bar-Hillel. 1960.
\newblock \href {https://doi.org/https://doi.org/10.1016/S0065-2458(08)60607-5}
  {The present status of automatic translation of languages**this article was
  prepared with the sponsorship of the informations systems branch, office of
  naval research, under contract nr 049130. reproduction as a whole or in part
  for the purposes of the u. s. government is permitted.}
\newblock volume~1 of \emph{Advances in Computers}, pages 91--163. Elsevier.

\bibitem[{Baziotis et~al.(2019)Baziotis, Androutsopoulos, Konstas, and
  Potamianos}]{baziotis-etal-2019-seq}
Christos Baziotis, Ion Androutsopoulos, Ioannis Konstas, and Alexandros
  Potamianos. 2019.
\newblock \href {https://doi.org/10.18653/v1/N19-1071} {{SEQ}{\^{}}3:
  Differentiable sequence-to-sequence-to-sequence autoencoder for unsupervised
  abstractive sentence compression}.
\newblock In \emph{Proceedings of the 2019 Conference of the North {A}merican
  Chapter of the Association for Computational Linguistics: Human Language
  Technologies, Volume 1 (Long and Short Papers)}, pages 673--681, Minneapolis,
  Minnesota. Association for Computational Linguistics.

\bibitem[{Bhagavatula et~al.(2020)Bhagavatula, Bras, Malaviya, Sakaguchi,
  Holtzman, Rashkin, Downey, tau Yih, and
  Choi}]{DBLP:conf/iclr/BhagavatulaBMSH20}
Chandra Bhagavatula, Ronan~Le Bras, Chaitanya Malaviya, Keisuke Sakaguchi, Ari
  Holtzman, Hannah Rashkin, Doug Downey, Wen tau Yih, and Yejin Choi. 2020.
\newblock \href {https://openreview.net/forum?id=Byg1v1HKDB} {Abductive
  commonsense reasoning.}
\newblock In \emph{ICLR}.

\bibitem[{Brown et~al.(2020)Brown, Mann, Ryder, Subbiah, Kaplan, Dhariwal,
  Neelakantan, Shyam, Sastry, Askell, Agarwal, Herbert-Voss, Krueger, Henighan,
  Child, Ramesh, Ziegler, Wu, Winter, Hesse, Chen, Sigler, Litwin, Gray, Chess,
  Clark, Berner, McCandlish, Radford, Sutskever, and
  Amodei}]{NEURIPS2020_1457c0d6}
Tom Brown, Benjamin Mann, Nick Ryder, Melanie Subbiah, Jared~D Kaplan, Prafulla
  Dhariwal, Arvind Neelakantan, Pranav Shyam, Girish Sastry, Amanda Askell,
  Sandhini Agarwal, Ariel Herbert-Voss, Gretchen Krueger, Tom Henighan, Rewon
  Child, Aditya Ramesh, Daniel Ziegler, Jeffrey Wu, Clemens Winter, Chris
  Hesse, Mark Chen, Eric Sigler, Mateusz Litwin, Scott Gray, Benjamin Chess,
  Jack Clark, Christopher Berner, Sam McCandlish, Alec Radford, Ilya Sutskever,
  and Dario Amodei. 2020.
\newblock \href
  {https://proceedings.neurips.cc/paper/2020/file/1457c0d6bfcb4967418bfb8ac142f64a-Paper.pdf}
  {Language models are few-shot learners}.
\newblock In \emph{Advances in Neural Information Processing Systems},
  volume~33, pages 1877--1901. Curran Associates, Inc.

\bibitem[{Carlsson et~al.(2022)Carlsson, {\"O}hman, Liu, Verlinden, Nivre, and
  Sahlgren}]{carlsson-etal-2022-fine}
Fredrik Carlsson, Joey {\"O}hman, Fangyu Liu, Severine Verlinden, Joakim Nivre,
  and Magnus Sahlgren. 2022.
\newblock \href {https://doi.org/10.18653/v1/2022.acl-long.471} {Fine-grained
  controllable text generation using non-residual prompting}.
\newblock In \emph{Proceedings of the 60th Annual Meeting of the Association
  for Computational Linguistics (Volume 1: Long Papers)}, pages 6837--6857,
  Dublin, Ireland. Association for Computational Linguistics.

\bibitem[{Chan et~al.(2021{\natexlab{a}})Chan, Ong, Pung, Zhang, and
  Fu}]{chan2021cocon}
Alvin Chan, Yew-Soon Ong, Bill Pung, Aston Zhang, and Jie Fu.
  2021{\natexlab{a}}.
\newblock \href {https://openreview.net/forum?id=VD_ozqvBy4W} {Cocon: A
  self-supervised approach for controlled text generation}.
\newblock In \emph{International Conference on Learning Representations}.

\bibitem[{Chan et~al.(2021{\natexlab{b}})Chan, Wang, and
  King}]{chan-etal-2021-controllable}
Hou~Pong Chan, Lu~Wang, and Irwin King. 2021{\natexlab{b}}.
\newblock \href {https://doi.org/10.1162/tacl_a_00423} {Controllable
  summarization with constrained {M}arkov decision process}.
\newblock \emph{Transactions of the Association for Computational Linguistics},
  9:1213--1232.

\bibitem[{Chung et~al.(2022)Chung, Hou, Longpre, Zoph, Tay, Fedus, Li, Wang,
  Dehghani, Brahma, Webson, Gu, Dai, Suzgun, Chen, Chowdhery, Narang, Mishra,
  Yu, Zhao, Huang, Dai, Yu, Petrov, Chi, Dean, Devlin, Roberts, Zhou, Le, and
  Wei}]{DBLP:journals/corr/abs-2210-11416}
Hyung~Won Chung, Le~Hou, Shayne Longpre, Barret Zoph, Yi~Tay, William Fedus,
  Eric Li, Xuezhi Wang, Mostafa Dehghani, Siddhartha Brahma, Albert Webson,
  Shixiang~Shane Gu, Zhuyun Dai, Mirac Suzgun, Xinyun Chen, Aakanksha
  Chowdhery, Sharan Narang, Gaurav Mishra, Adams Yu, Vincent~Y. Zhao, Yanping
  Huang, Andrew~M. Dai, Hongkun Yu, Slav Petrov, Ed~H. Chi, Jeff Dean, Jacob
  Devlin, Adam Roberts, Denny Zhou, Quoc~V. Le, and Jason Wei. 2022.
\newblock \href {https://doi.org/10.48550/arXiv.2210.11416} {Scaling
  instruction-finetuned language models}.
\newblock \emph{CoRR}, abs/2210.11416.

\bibitem[{Devlin et~al.(2019)Devlin, Chang, Lee, and
  Toutanova}]{devlin-etal-2019-bert}
Jacob Devlin, Ming-Wei Chang, Kenton Lee, and Kristina Toutanova. 2019.
\newblock \href {https://doi.org/10.18653/v1/N19-1423} {{BERT}: Pre-training of
  deep bidirectional transformers for language understanding}.
\newblock In \emph{Proceedings of the 2019 Conference of the North {A}merican
  Chapter of the Association for Computational Linguistics: Human Language
  Technologies, Volume 1 (Long and Short Papers)}, pages 4171--4186,
  Minneapolis, Minnesota. Association for Computational Linguistics.

\bibitem[{Dinu et~al.(2019)Dinu, Mathur, Federico, and
  Al-Onaizan}]{dinu-etal-2019-training}
Georgiana Dinu, Prashant Mathur, Marcello Federico, and Yaser Al-Onaizan. 2019.
\newblock \href {https://doi.org/10.18653/v1/P19-1294} {Training neural machine
  translation to apply terminology constraints}.
\newblock In \emph{Proceedings of the 57th Annual Meeting of the Association
  for Computational Linguistics}, pages 3063--3068, Florence, Italy.
  Association for Computational Linguistics.

\bibitem[{Fan et~al.(2018)Fan, Grangier, and Auli}]{fan-etal-2018-controllable}
Angela Fan, David Grangier, and Michael Auli. 2018.
\newblock \href {https://doi.org/10.18653/v1/W18-2706} {Controllable
  abstractive summarization}.
\newblock In \emph{Proceedings of the 2nd Workshop on Neural Machine
  Translation and Generation}, pages 45--54, Melbourne, Australia. Association
  for Computational Linguistics.

\bibitem[{Graff et~al.(2003)Graff, Kong, Chen, and Maeda}]{graff2003english}
David Graff, Junbo Kong, Ke~Chen, and Kazuaki Maeda. 2003.
\newblock English gigaword.
\newblock \emph{Linguistic Data Consortium, Philadelphia}, 4(1):34.

\bibitem[{He et~al.(2022)He, Kryscinski, McCann, Rajani, and
  Xiong}]{he-etal-2022-ctrlsum}
Junxian He, Wojciech Kryscinski, Bryan McCann, Nazneen Rajani, and Caiming
  Xiong. 2022.
\newblock \href {https://doi.org/10.18653/v1/2022.emnlp-main.396} {{CTRL}sum:
  Towards generic controllable text summarization}.
\newblock In \emph{Proceedings of the 2022 Conference on Empirical Methods in
  Natural Language Processing}, pages 5879--5915, Abu Dhabi, United Arab
  Emirates. Association for Computational Linguistics.

\bibitem[{He(2021)}]{he-2021-parallel}
Xingwei He. 2021.
\newblock \href {https://doi.org/10.18653/v1/2021.emnlp-main.681} {Parallel
  refinements for lexically constrained text generation with {BART}}.
\newblock In \emph{Proceedings of the 2021 Conference on Empirical Methods in
  Natural Language Processing}, pages 8653--8666, Online and Punta Cana,
  Dominican Republic. Association for Computational Linguistics.

\bibitem[{Keskar et~al.(2019)Keskar, McCann, Varshney, Xiong, and
  Socher}]{DBLP:journals/corr/abs-1909-05858}
Nitish~Shirish Keskar, Bryan McCann, Lav~R. Varshney, Caiming Xiong, and
  Richard Socher. 2019.
\newblock \href {http://arxiv.org/abs/1909.05858} {{CTRL:} {A} conditional
  transformer language model for controllable generation}.
\newblock \emph{CoRR}, abs/1909.05858.

\bibitem[{Khashabi et~al.(2021)Khashabi, Stanovsky, Bragg, Lourie, Kasai, Choi,
  Smith, and Weld}]{khashabi2021genie}
Daniel Khashabi, Gabriel Stanovsky, Jonathan Bragg, Nicholas Lourie, Jungo
  Kasai, Yejin Choi, Noah~A Smith, and Daniel~S Weld. 2021.
\newblock Genie: A leaderboard for human-in-the-loop evaluation of text
  generation.
\newblock \emph{arXiv preprint arXiv:2101.06561}.

\bibitem[{Kikuchi et~al.(2016)Kikuchi, Neubig, Sasano, Takamura, and
  Okumura}]{kikuchi-etal-2016-controlling}
Yuta Kikuchi, Graham Neubig, Ryohei Sasano, Hiroya Takamura, and Manabu
  Okumura. 2016.
\newblock \href {https://doi.org/10.18653/v1/D16-1140} {Controlling output
  length in neural encoder-decoders}.
\newblock In \emph{Proceedings of the 2016 Conference on Empirical Methods in
  Natural Language Processing}, pages 1328--1338, Austin, Texas. Association
  for Computational Linguistics.

\bibitem[{Kojima et~al.(2022)Kojima, Gu, Reid, Matsuo, and
  Iwasawa}]{kojima2022large}
Takeshi Kojima, Shixiang~Shane Gu, Machel Reid, Yutaka Matsuo, and Yusuke
  Iwasawa. 2022.
\newblock \href {https://openreview.net/forum?id=e2TBb5y0yFf} {Large language
  models are zero-shot reasoners}.
\newblock In \emph{Advances in Neural Information Processing Systems}.

\bibitem[{Krause et~al.(2021)Krause, Gotmare, McCann, Keskar, Joty, Socher, and
  Rajani}]{krause-etal-2021-gedi-generative}
Ben Krause, Akhilesh~Deepak Gotmare, Bryan McCann, Nitish~Shirish Keskar,
  Shafiq Joty, Richard Socher, and Nazneen~Fatema Rajani. 2021.
\newblock \href {https://doi.org/10.18653/v1/2021.findings-emnlp.424}
  {{G}e{D}i: Generative discriminator guided sequence generation}.
\newblock In \emph{Findings of the Association for Computational Linguistics:
  EMNLP 2021}, pages 4929--4952, Punta Cana, Dominican Republic. Association
  for Computational Linguistics.

\bibitem[{Lai et~al.(2017)Lai, Xie, Liu, Yang, and Hovy}]{lai-etal-2017-race}
Guokun Lai, Qizhe Xie, Hanxiao Liu, Yiming Yang, and Eduard Hovy. 2017.
\newblock \href {https://doi.org/10.18653/v1/D17-1082} {{RACE}: Large-scale
  {R}e{A}ding comprehension dataset from examinations}.
\newblock In \emph{Proceedings of the 2017 Conference on Empirical Methods in
  Natural Language Processing}, pages 785--794, Copenhagen, Denmark.
  Association for Computational Linguistics.

\bibitem[{Lester et~al.(2021)Lester, Al-Rfou, and
  Constant}]{lester-etal-2021-power}
Brian Lester, Rami Al-Rfou, and Noah Constant. 2021.
\newblock \href {https://doi.org/10.18653/v1/2021.emnlp-main.243} {The power of
  scale for parameter-efficient prompt tuning}.
\newblock In \emph{Proceedings of the 2021 Conference on Empirical Methods in
  Natural Language Processing}, pages 3045--3059, Online and Punta Cana,
  Dominican Republic. Association for Computational Linguistics.

\bibitem[{Lewis et~al.(2020)Lewis, Liu, Goyal, Ghazvininejad, Mohamed, Levy,
  Stoyanov, and Zettlemoyer}]{lewis-etal-2020-bart}
Mike Lewis, Yinhan Liu, Naman Goyal, Marjan Ghazvininejad, Abdelrahman Mohamed,
  Omer Levy, Veselin Stoyanov, and Luke Zettlemoyer. 2020.
\newblock \href {https://doi.org/10.18653/v1/2020.acl-main.703} {{BART}:
  Denoising sequence-to-sequence pre-training for natural language generation,
  translation, and comprehension}.
\newblock In \emph{Proceedings of the 58th Annual Meeting of the Association
  for Computational Linguistics}, pages 7871--7880, Online. Association for
  Computational Linguistics.

\bibitem[{Li et~al.(2020)Li, Zhang, Liu, and Shi}]{li-etal-2020-rigid}
Piji Li, Haisong Zhang, Xiaojiang Liu, and Shuming Shi. 2020.
\newblock \href {https://doi.org/10.18653/v1/2020.acl-main.68} {Rigid formats
  controlled text generation}.
\newblock In \emph{Proceedings of the 58th Annual Meeting of the Association
  for Computational Linguistics}, pages 742--751, Online. Association for
  Computational Linguistics.

\bibitem[{Li et~al.(2022)Li, Thickstun, Gulrajani, Liang, and
  Hashimoto}]{li2022diffusionlm}
Xiang~Lisa Li, John Thickstun, Ishaan Gulrajani, Percy Liang, and Tatsunori
  Hashimoto. 2022.
\newblock \href {https://openreview.net/forum?id=3s9IrEsjLyk} {Diffusion-{LM}
  improves controllable text generation}.
\newblock In \emph{Advances in Neural Information Processing Systems}.

\bibitem[{Lin et~al.(2020)Lin, Zhou, Shen, Zhou, Bhagavatula, Choi, and
  Ren}]{lin-etal-2020-commongen}
Bill~Yuchen Lin, Wangchunshu Zhou, Ming Shen, Pei Zhou, Chandra Bhagavatula,
  Yejin Choi, and Xiang Ren. 2020.
\newblock \href {https://doi.org/10.18653/v1/2020.findings-emnlp.165}
  {{C}ommon{G}en: A constrained text generation challenge for generative
  commonsense reasoning}.
\newblock In \emph{Findings of the Association for Computational Linguistics:
  EMNLP 2020}, pages 1823--1840, Online. Association for Computational
  Linguistics.

\bibitem[{Lin(2004)}]{lin-2004-rouge}
Chin-Yew Lin. 2004.
\newblock \href {https://aclanthology.org/W04-1013} {{ROUGE}: A package for
  automatic evaluation of summaries}.
\newblock In \emph{Text Summarization Branches Out}, pages 74--81, Barcelona,
  Spain. Association for Computational Linguistics.

\bibitem[{Liu et~al.(2022)Liu, Su, Shareghi, and Collier}]{liu-etal-2022-plug}
Yinhong Liu, Yixuan Su, Ehsan Shareghi, and Nigel Collier. 2022.
\newblock \href {https://aclanthology.org/2022.gem-1.19} {Plug-and-play recipe
  generation with content planning}.
\newblock In \emph{Proceedings of the 2nd Workshop on Natural Language
  Generation, Evaluation, and Metrics (GEM)}, pages 223--234, Abu Dhabi, United
  Arab Emirates (Hybrid). Association for Computational Linguistics.

\bibitem[{Logeswaran et~al.(2018)Logeswaran, Lee, and
  Bengio}]{10.5555/3327345.3327417}
Lajanugen Logeswaran, Honglak Lee, and Samy Bengio. 2018.
\newblock Content preserving text generation with attribute controls.
\newblock In \emph{Proceedings of the 32nd International Conference on Neural
  Information Processing Systems}, NIPS'18, page 5108–5118, Red Hook, NY,
  USA. Curran Associates Inc.

\bibitem[{Lu et~al.(2022)Lu, Welleck, West, Jiang, Kasai, Khashabi, Le~Bras,
  Qin, Yu, Zellers, Smith, and Choi}]{lu-etal-2022-neurologic}
Ximing Lu, Sean Welleck, Peter West, Liwei Jiang, Jungo Kasai, Daniel Khashabi,
  Ronan Le~Bras, Lianhui Qin, Youngjae Yu, Rowan Zellers, Noah~A. Smith, and
  Yejin Choi. 2022.
\newblock \href {https://doi.org/10.18653/v1/2022.naacl-main.57}
  {{N}euro{L}ogic a*esque decoding: Constrained text generation with lookahead
  heuristics}.
\newblock In \emph{Proceedings of the 2022 Conference of the North American
  Chapter of the Association for Computational Linguistics: Human Language
  Technologies}, pages 780--799, Seattle, United States. Association for
  Computational Linguistics.

\bibitem[{Lu et~al.(2021)Lu, West, Zellers, Le~Bras, Bhagavatula, and
  Choi}]{lu-etal-2021-neurologic}
Ximing Lu, Peter West, Rowan Zellers, Ronan Le~Bras, Chandra Bhagavatula, and
  Yejin Choi. 2021.
\newblock \href {https://doi.org/10.18653/v1/2021.naacl-main.339}
  {{N}euro{L}ogic decoding: (un)supervised neural text generation with
  predicate logic constraints}.
\newblock In \emph{Proceedings of the 2021 Conference of the North American
  Chapter of the Association for Computational Linguistics: Human Language
  Technologies}, pages 4288--4299, Online. Association for Computational
  Linguistics.

\bibitem[{Luhn(1957)}]{5392697}
H.~P. Luhn. 1957.
\newblock \href {https://doi.org/10.1147/rd.14.0309} {A statistical approach to
  mechanized encoding and searching of literary information}.
\newblock \emph{IBM Journal of Research and Development}, 1(4):309--317.

\bibitem[{Madaan and Yazdanbakhsh(2022)}]{DBLP:journals/corr/abs-2209-07686}
Aman Madaan and Amir Yazdanbakhsh. 2022.
\newblock \href {https://doi.org/10.48550/arXiv.2209.07686} {Text and patterns:
  For effective chain of thought, it takes two to tango}.
\newblock \emph{CoRR}, abs/2209.07686.

\bibitem[{Meng et~al.(2022)Meng, Lu, Peng, and Chang}]{meng2022controllable}
Tao Meng, Sidi Lu, Nanyun Peng, and Kai-Wei Chang. 2022.
\newblock \href {https://openreview.net/forum?id=yI7i9yc3Upr} {Controllable
  text generation with neurally-decomposed oracle}.
\newblock In \emph{Advances in Neural Information Processing Systems}.

\bibitem[{Min et~al.(2022)Min, Lyu, Holtzman, Artetxe, Lewis, Hajishirzi, and
  Zettlemoyer}]{DBLP:journals/corr/abs-2202-12837}
Sewon Min, Xinxi Lyu, Ari Holtzman, Mikel Artetxe, Mike Lewis, Hannaneh
  Hajishirzi, and Luke Zettlemoyer. 2022.
\newblock \href {http://arxiv.org/abs/2202.12837} {Rethinking the role of
  demonstrations: What makes in-context learning work?}
\newblock \emph{CoRR}, abs/2202.12837.

\bibitem[{Mostafazadeh et~al.(2016)Mostafazadeh, Chambers, He, Parikh, Batra,
  Vanderwende, Kohli, and Allen}]{mostafazadeh-etal-2016-corpus}
Nasrin Mostafazadeh, Nathanael Chambers, Xiaodong He, Devi Parikh, Dhruv Batra,
  Lucy Vanderwende, Pushmeet Kohli, and James Allen. 2016.
\newblock \href {https://doi.org/10.18653/v1/N16-1098} {A corpus and cloze
  evaluation for deeper understanding of commonsense stories}.
\newblock In \emph{Proceedings of the 2016 Conference of the North {A}merican
  Chapter of the Association for Computational Linguistics: Human Language
  Technologies}, pages 839--849, San Diego, California. Association for
  Computational Linguistics.

\bibitem[{Novikova et~al.(2017)Novikova, Du{\v{s}}ek, and
  Rieser}]{novikova-etal-2017-e2e}
Jekaterina Novikova, Ond{\v{r}}ej Du{\v{s}}ek, and Verena Rieser. 2017.
\newblock \href {https://doi.org/10.18653/v1/W17-5525} {The {E}2{E} dataset:
  New challenges for end-to-end generation}.
\newblock In \emph{Proceedings of the 18th Annual {SIG}dial Meeting on
  Discourse and Dialogue}, pages 201--206, Saarbr{\"u}cken, Germany.
  Association for Computational Linguistics.

\bibitem[{Ouyang et~al.(2022)Ouyang, Wu, Jiang, Almeida, Wainwright, Mishkin,
  Zhang, Agarwal, Slama, Ray, Schulman, Hilton, Kelton, Miller, Simens, Askell,
  Welinder, Christiano, Leike, and Lowe}]{DBLP:journals/corr/abs-2203-02155}
Long Ouyang, Jeff Wu, Xu~Jiang, Diogo Almeida, Carroll~L. Wainwright, Pamela
  Mishkin, Chong Zhang, Sandhini Agarwal, Katarina Slama, Alex Ray, John
  Schulman, Jacob Hilton, Fraser Kelton, Luke Miller, Maddie Simens, Amanda
  Askell, Peter Welinder, Paul~F. Christiano, Jan Leike, and Ryan Lowe. 2022.
\newblock \href {https://doi.org/10.48550/arXiv.2203.02155} {Training language
  models to follow instructions with human feedback}.
\newblock \emph{CoRR}, abs/2203.02155.

\bibitem[{Papineni et~al.(2002)Papineni, Roukos, Ward, and
  Zhu}]{papineni-etal-2002-bleu}
Kishore Papineni, Salim Roukos, Todd Ward, and Wei-Jing Zhu. 2002.
\newblock \href {https://doi.org/10.3115/1073083.1073135} {{B}leu: a method for
  automatic evaluation of machine translation}.
\newblock In \emph{Proceedings of the 40th Annual Meeting of the Association
  for Computational Linguistics}, pages 311--318, Philadelphia, Pennsylvania,
  USA. Association for Computational Linguistics.

\bibitem[{Prabhumoye et~al.(2020)Prabhumoye, Black, and
  Salakhutdinov}]{prabhumoye-etal-2020-exploring}
Shrimai Prabhumoye, Alan~W Black, and Ruslan Salakhutdinov. 2020.
\newblock \href {https://doi.org/10.18653/v1/2020.coling-main.1} {Exploring
  controllable text generation techniques}.
\newblock In \emph{Proceedings of the 28th International Conference on
  Computational Linguistics}, pages 1--14, Barcelona, Spain (Online).
  International Committee on Computational Linguistics.

\bibitem[{Qin et~al.(2020)Qin, Shwartz, West, Bhagavatula, Hwang, Le~Bras,
  Bosselut, and Choi}]{qin-etal-2020-back}
Lianhui Qin, Vered Shwartz, Peter West, Chandra Bhagavatula, Jena~D. Hwang,
  Ronan Le~Bras, Antoine Bosselut, and Yejin Choi. 2020.
\newblock \href {https://doi.org/10.18653/v1/2020.emnlp-main.58} {Back to the
  future: Unsupervised backprop-based decoding for counterfactual and abductive
  commonsense reasoning}.
\newblock In \emph{Proceedings of the 2020 Conference on Empirical Methods in
  Natural Language Processing (EMNLP)}, pages 794--805, Online. Association for
  Computational Linguistics.

\bibitem[{Qin et~al.(2022)Qin, Welleck, Khashabi, and Choi}]{qin2022cold}
Lianhui Qin, Sean Welleck, Daniel Khashabi, and Yejin Choi. 2022.
\newblock \href {https://openreview.net/forum?id=TiZYrQ-mPup} {{COLD} decoding:
  Energy-based constrained text generation with langevin dynamics}.
\newblock In \emph{Advances in Neural Information Processing Systems}.

\bibitem[{Raffel et~al.(2022)Raffel, Shazeer, Roberts, Lee, Narang, Matena,
  Zhou, Li, and Liu}]{10.5555/3455716.3455856}
Colin Raffel, Noam Shazeer, Adam Roberts, Katherine Lee, Sharan Narang, Michael
  Matena, Yanqi Zhou, Wei Li, and Peter~J. Liu. 2022.
\newblock Exploring the limits of transfer learning with a unified text-to-text
  transformer.
\newblock \emph{J. Mach. Learn. Res.}, 21(1).

\bibitem[{Rosenbaum et~al.(2022)Rosenbaum, Soltan, Hamza, Versley, and
  Boese}]{rosenbaum-etal-2022-linguist}
Andy Rosenbaum, Saleh Soltan, Wael Hamza, Yannick Versley, and Markus Boese.
  2022.
\newblock \href {https://aclanthology.org/2022.coling-1.18} {{LINGUIST}:
  Language model instruction tuning to generate annotated utterances for intent
  classification and slot tagging}.
\newblock In \emph{Proceedings of the 29th International Conference on
  Computational Linguistics}, pages 218--241, Gyeongju, Republic of Korea.
  International Committee on Computational Linguistics.

\bibitem[{Shen et~al.(2020)Shen, Quach, Barzilay, and
  Jaakkola}]{shen-etal-2020-blank}
Tianxiao Shen, Victor Quach, Regina Barzilay, and Tommi Jaakkola. 2020.
\newblock \href {https://doi.org/10.18653/v1/2020.emnlp-main.420} {Blank
  language models}.
\newblock In \emph{Proceedings of the 2020 Conference on Empirical Methods in
  Natural Language Processing (EMNLP)}, pages 5186--5198, Online. Association
  for Computational Linguistics.

\bibitem[{Stiennon et~al.(2020{\natexlab{a}})Stiennon, Ouyang, Wu, Ziegler,
  Lowe, Voss, Radford, Amodei, and Christiano}]{10.5555/3495724.3495977}
Nisan Stiennon, Long Ouyang, Jeff Wu, Daniel~M. Ziegler, Ryan Lowe, Chelsea
  Voss, Alec Radford, Dario Amodei, and Paul Christiano. 2020{\natexlab{a}}.
\newblock Learning to summarize from human feedback.
\newblock In \emph{Proceedings of the 34th International Conference on Neural
  Information Processing Systems}, NIPS'20, Red Hook, NY, USA. Curran
  Associates Inc.

\bibitem[{Stiennon et~al.(2020{\natexlab{b}})Stiennon, Ouyang, Wu, Ziegler,
  Lowe, Voss, Radford, Amodei, and Christiano}]{NEURIPS2020_1f89885d}
Nisan Stiennon, Long Ouyang, Jeffrey Wu, Daniel Ziegler, Ryan Lowe, Chelsea
  Voss, Alec Radford, Dario Amodei, and Paul~F Christiano. 2020{\natexlab{b}}.
\newblock \href
  {https://proceedings.neurips.cc/paper/2020/file/1f89885d556929e98d3ef9b86448f951-Paper.pdf}
  {Learning to summarize with human feedback}.
\newblock In \emph{Advances in Neural Information Processing Systems},
  volume~33, pages 3008--3021. Curran Associates, Inc.

\bibitem[{Su et~al.(2021)Su, Vandyke, Wang, Fang, and
  Collier}]{su-etal-2021-plan-generate}
Yixuan Su, David Vandyke, Sihui Wang, Yimai Fang, and Nigel Collier. 2021.
\newblock \href {https://doi.org/10.18653/v1/2021.findings-emnlp.76}
  {Plan-then-generate: Controlled data-to-text generation via planning}.
\newblock In \emph{Findings of the Association for Computational Linguistics:
  EMNLP 2021}, pages 895--909, Punta Cana, Dominican Republic. Association for
  Computational Linguistics.

\bibitem[{Sun et~al.(2022)Sun, Jiang, Huang, Cao, Cheng, and
  Wang}]{DBLP:journals/corr/abs-2209-11409}
Zewei Sun, Qingnan Jiang, Shujian Huang, Jun Cao, Shanbo Cheng, and Mingxuan
  Wang. 2022.
\newblock \href {https://doi.org/10.48550/arXiv.2209.11409} {Zero-shot domain
  adaptation for neural machine translation with retrieved phrase-level
  prompts}.
\newblock \emph{CoRR}, abs/2209.11409.

\bibitem[{Susanto et~al.(2020)Susanto, Chollampatt, and
  Tan}]{susanto-etal-2020-lexically}
Raymond~Hendy Susanto, Shamil Chollampatt, and Liling Tan. 2020.
\newblock \href {https://doi.org/10.18653/v1/2020.acl-main.325} {Lexically
  constrained neural machine translation with {L}evenshtein transformer}.
\newblock In \emph{Proceedings of the 58th Annual Meeting of the Association
  for Computational Linguistics}, pages 3536--3543, Online. Association for
  Computational Linguistics.

\bibitem[{Tang et~al.(2019)Tang, Li, and Jin}]{tang-etal-2019-topic}
Hongyin Tang, Miao Li, and Beihong Jin. 2019.
\newblock \href {https://doi.org/10.18653/v1/D19-1513} {A topic augmented text
  generation model: Joint learning of semantics and structural features}.
\newblock In \emph{Proceedings of the 2019 Conference on Empirical Methods in
  Natural Language Processing and the 9th International Joint Conference on
  Natural Language Processing (EMNLP-IJCNLP)}, pages 5090--5099, Hong Kong,
  China. Association for Computational Linguistics.

\bibitem[{Tay et~al.(2022)Tay, Dehghani, Tran, Garcia, Bahri, Schuster, Zheng,
  Houlsby, and Metzler}]{DBLP:journals/corr/abs-2205-05131}
Yi~Tay, Mostafa Dehghani, Vinh~Q. Tran, Xavier Garcia, Dara Bahri, Tal
  Schuster, Huaixiu~Steven Zheng, Neil Houlsby, and Donald Metzler. 2022.
\newblock \href {https://doi.org/10.48550/arXiv.2205.05131} {Unifying language
  learning paradigms}.
\newblock \emph{CoRR}, abs/2205.05131.

\bibitem[{Vedantam et~al.(2015)Vedantam, Lawrence~Zitnick, and
  Parikh}]{Vedantam_2015_CVPR}
Ramakrishna Vedantam, C.~Lawrence~Zitnick, and Devi Parikh. 2015.
\newblock Cider: Consensus-based image description evaluation.
\newblock In \emph{Proceedings of the IEEE Conference on Computer Vision and
  Pattern Recognition (CVPR)}.

\bibitem[{Wang et~al.(2021)Wang, Wood, Wan, Dras, and
  Johnson}]{wang-etal-2021-mention}
Yufei Wang, Ian Wood, Stephen Wan, Mark Dras, and Mark Johnson. 2021.
\newblock \href {https://doi.org/10.18653/v1/2021.acl-long.9} {Mention flags
  ({MF}): Constraining transformer-based text generators}.
\newblock In \emph{Proceedings of the 59th Annual Meeting of the Association
  for Computational Linguistics and the 11th International Joint Conference on
  Natural Language Processing (Volume 1: Long Papers)}, pages 103--113, Online.
  Association for Computational Linguistics.

\bibitem[{Wolf et~al.(2020)Wolf, Debut, Sanh, Chaumond, Delangue, Moi, Cistac,
  Rault, Louf, Funtowicz, Davison, Shleifer, von Platen, Ma, Jernite, Plu, Xu,
  Le~Scao, Gugger, Drame, Lhoest, and Rush}]{wolf-etal-2020-transformers}
Thomas Wolf, Lysandre Debut, Victor Sanh, Julien Chaumond, Clement Delangue,
  Anthony Moi, Pierric Cistac, Tim Rault, Remi Louf, Morgan Funtowicz, Joe
  Davison, Sam Shleifer, Patrick von Platen, Clara Ma, Yacine Jernite, Julien
  Plu, Canwen Xu, Teven Le~Scao, Sylvain Gugger, Mariama Drame, Quentin Lhoest,
  and Alexander Rush. 2020.
\newblock \href {https://doi.org/10.18653/v1/2020.emnlp-demos.6} {Transformers:
  State-of-the-art natural language processing}.
\newblock In \emph{Proceedings of the 2020 Conference on Empirical Methods in
  Natural Language Processing: System Demonstrations}, pages 38--45, Online.
  Association for Computational Linguistics.

\bibitem[{Xu and Carpuat(2021)}]{10.1162/tacl_a_00368}
Weijia Xu and Marine Carpuat. 2021.
\newblock \href {https://doi.org/10.1162/tacl_a_00368} {{EDITOR: An Edit-Based
  Transformer with Repositioning for Neural Machine Translation with Soft
  Lexical Constraints}}.
\newblock \emph{Transactions of the Association for Computational Linguistics},
  9:311--328.

\bibitem[{Yang and Klein(2021)}]{yang-klein-2021-fudge}
Kevin Yang and Dan Klein. 2021.
\newblock \href {https://doi.org/10.18653/v1/2021.naacl-main.276} {{FUDGE}:
  Controlled text generation with future discriminators}.
\newblock In \emph{Proceedings of the 2021 Conference of the North American
  Chapter of the Association for Computational Linguistics: Human Language
  Technologies}, pages 3511--3535, Online. Association for Computational
  Linguistics.

\bibitem[{Yang et~al.(2020)Yang, Zhu, Gmyr, Zeng, Huang, and
  Darve}]{yang-etal-2020-ted}
Ziyi Yang, Chenguang Zhu, Robert Gmyr, Michael Zeng, Xuedong Huang, and Eric
  Darve. 2020.
\newblock \href {https://doi.org/10.18653/v1/2020.findings-emnlp.168} {{TED}: A
  pretrained unsupervised summarization model with theme modeling and
  denoising}.
\newblock In \emph{Findings of the Association for Computational Linguistics:
  EMNLP 2020}, pages 1865--1874, Online. Association for Computational
  Linguistics.

\bibitem[{Zhang et~al.(2022)Zhang, Song, Li, Zhou, and
  Song}]{DBLP:journals/corr/abs-2201-05337}
Hanqing Zhang, Haolin Song, Shaoyu Li, Ming Zhou, and Dawei Song. 2022.
\newblock \href {http://arxiv.org/abs/2201.05337} {A survey of controllable
  text generation using transformer-based pre-trained language models}.
\newblock \emph{CoRR}, abs/2201.05337.

\bibitem[{Zhang* et~al.(2020)Zhang*, Kishore*, Wu*, Weinberger, and
  Artzi}]{Zhang*2020BERTScore:}
Tianyi Zhang*, Varsha Kishore*, Felix Wu*, Kilian~Q. Weinberger, and Yoav
  Artzi. 2020.
\newblock \href {https://openreview.net/forum?id=SkeHuCVFDr} {Bertscore:
  Evaluating text generation with bert}.
\newblock In \emph{International Conference on Learning Representations}.

\bibitem[{Zhang et~al.(2020)Zhang, Wang, Li, Gan, Brockett, and
  Dolan}]{zhang-etal-2020-pointer}
Yizhe Zhang, Guoyin Wang, Chunyuan Li, Zhe Gan, Chris Brockett, and Bill Dolan.
  2020.
\newblock \href {https://doi.org/10.18653/v1/2020.emnlp-main.698} {{POINTER}:
  Constrained progressive text generation via insertion-based generative
  pre-training}.
\newblock In \emph{Proceedings of the 2020 Conference on Empirical Methods in
  Natural Language Processing (EMNLP)}, pages 8649--8670, Online. Association
  for Computational Linguistics.

\bibitem[{Zhang et~al.(2023)Zhang, Liu, Yang, Fang, Chen, Radev, Zhu, Zeng, and
  Zhang}]{10.1162/tacl_a_00575}
Yusen Zhang, Yang Liu, Ziyi Yang, Yuwei Fang, Yulong Chen, Dragomir Radev,
  Chenguang Zhu, Michael Zeng, and Rui Zhang. 2023.
\newblock \href {https://doi.org/10.1162/tacl_a_00575} {{MACSum: Controllable
  Summarization with Mixed Attributes}}.
\newblock \emph{Transactions of the Association for Computational Linguistics},
  11:787--803.

\bibitem[{Zhou et~al.(2023)Zhou, Jiang, Wilcox, Cotterell, and
  Sachan}]{DBLP:journals/corr/abs-2304-14293}
Wangchunshu Zhou, Yuchen~Eleanor Jiang, Ethan Wilcox, Ryan Cotterell, and
  Mrinmaya Sachan. 2023.
\newblock \href {https://doi.org/10.48550/arXiv.2304.14293} {Controlled text
  generation with natural language instructions}.
\newblock \emph{CoRR}, abs/2304.14293.

\bibitem[{Ziegler et~al.(2019)Ziegler, Stiennon, Wu, Brown, Radford, Amodei,
  Christiano, and Irving}]{DBLP:journals/corr/abs-1909-08593}
Daniel~M. Ziegler, Nisan Stiennon, Jeffrey Wu, Tom~B. Brown, Alec Radford,
  Dario Amodei, Paul~F. Christiano, and Geoffrey Irving. 2019.
\newblock \href {http://arxiv.org/abs/1909.08593} {Fine-tuning language models
  from human preferences}.
\newblock \emph{CoRR}, abs/1909.08593.

\end{thebibliography}
\bibliographystyle{acl_natbib}

\appendix

\section{Recursive Algorithm}
\begin{algorithm}
\caption{Recursive decoding}
\label{alg:recursive_decoding}
\begin{algorithmic}[1]
\Function{RecursiveDecoding}{exp}
\If {not ContainNonterminal(exp)}
    \State \Return input
\EndIf
\If {not ContainOptions(exp)}
    \State \Return Generate(exp)
\EndIf  

\State {exp\_before\_opts, opts, exp\_after\_opts $\gets$ SplitByFirstOption(exp)}

\For {i, ch in enumerate(opts)}
    \State exp\_ch $\gets$ exp\_before\_opts+ch
    \State opts[i] $\gets$ Genereate(exp\_ch)
\EndFor
\State best\_ch $\gets$ Generate(opts)
\State remain\_res $\gets$ Generate(exp\_after\_opts)
\Return best\_ch + remain\_res
\EndFunction
\end{algorithmic}
\end{algorithm}

\section{Dataset Statistics}
\begin{table}[h]
\begin{tabular}{llll}
\toprule
\textbf{Dataset} & \textbf{Train} & \textbf{Validation} & \textbf{Test} \\
\midrule
$\alpha$NLG & 50481 & 1780 & 3561 \\
$\alpha$NLI & 169654 & 1532 & 3059 \\
CommonGen & 67216 & 993 & - \\
Gigaword & - & 189644 & 1933 \\
IATE & - & - & 414 \\
Wiktionary & - & - & 727 \\
StoryCompletion & - & 1871 & 1871 \\
\bottomrule
\end{tabular}
\caption{Staststics of used dataset}
\end{table}

\section{Generation Statistics}
The data of average try and the first-time success rate during generation is presented in Table \ref{table:generation_statstics}. REI models tend to succeed on the first attempt for simple constraints, and only for challenging constraints, REI models would retry. Also, fine-tuned FLAN needs the least retry, while natural language instruction requires the most retry and may not be likely to succeed on the first try.
\begin{table*}[ht!]
\small
\centering
\begin{tabular}{@{}lllllll@{}}
\toprule
\multicolumn{1}{c}{\multirow{3}{*}{\textbf{Task}}} & \multicolumn{6}{c}{\textbf{Method}} \\
\multicolumn{1}{c}{} & \multicolumn{2}{c}{REI+FLAN-T5-xl} & \multicolumn{2}{c}{REI+GPT-3.5, 8 shot} & \multicolumn{2}{c}{NLI+GPT-3.5, 8 shot} \\
\multicolumn{1}{c}{} & \multicolumn{1}{c}{\textbf{Avg. Try}} & \multicolumn{1}{c}{\textbf{First SR.}} & \multicolumn{1}{c}{\textbf{Avg. Try}} & \multicolumn{1}{c}{\textbf{First SR.}} & \multicolumn{1}{c}{\textbf{Avg. Try}} & \multicolumn{1}{c}{\textbf{First SR.}} \\ \midrule
aNLG, length & 1.00 & 99.9 & 2.45 & 46.5 & 4.00 & 17.8 \\
aNLG, lexicon & 1.68 & 63.7 & 1.08 & 98.9 & 1.11 & 98.0 \\
aNLG, length \& lexicon & 1.39 & 76.6 & 2.57 & 52.5 & 4.29 & 10.0 \\
CommonGen & 1.01 & 98.7 & 1.39 & 86.1 & 1.95 & 67.1 \\
CommonGen, length & 1.04 & 96.9 & 2.23 & 60.4 & 4.07 & 17.9 \\
Gigaword, length & 1.05 & 94.9 & 3.48 & 35.6 & 4.07 & 14.4 \\
IATE & - & - & 1.01 & 99.0 & 1.19 & 91.9 \\
Wiktionary & - & - & 1.05 & 97.0 & 1.08 & 92.8 \\
StoryCloze, position & 1.00 & 100.0 & 1.04 & 97.0 & 1.01 & 78.9 \\ \bottomrule
\end{tabular}
\caption{Staststics of generation, presenting the average try (Avg. Try) and the first-time success rate (First SR.).}
\label{table:generation_statstics}
\end{table*}

\section{Natural Language Instruction Examples}
For the method of Natural Language Instruction on GPT-3.5, the instructions used on each task are shown in Table \ref{table:nli_example}.

\begin{table*}[]
\small
\centering
\begin{tabular}{lp{11.5cm}}
\toprule
\multicolumn{1}{c}{\textbf{Task}} & \multicolumn{1}{c}{\textbf{Instruction Example}} \\ \midrule
aNLG & The first sentence is " The Smiths were having holidays done of the children. " and the last sentence is " Ty's face lit up as he ran to the new toy, happily posing for photos. " . Insert a middle sentence with similar style, and the length shall not exceed 10 words. \\
aNLG, length & The first sentence is " The Smiths were having holidays done of the children. " and the last sentence is " Ty's face lit up as he ran to the new toy, happily posing for photos. " . Insert a middle sentence with similar style, and the length shall be exactly 7 words without counting punctuation. \\
aNLG, lexicon & The first sentence is " The Smiths were having holidays done of the children. " and the last sentence is " Ty's face lit up as he ran to the new toy, happily posing for photos. " . Insert a middle sentence with similar style, while also containing the keyword "bought". \\
aNLG, length \& lexicon & The first sentence is " The Smiths were having holidays done of the children. " and the last sentence is " Ty's face lit up as he ran to the new toy, happily posing for photos. " . Insert a middle sentence with similar style, while also containing the keyword "bought", and the length shall be exactly 7 words without counting punctuation. \\
CommonGen & Generate a sentence that mentions all of these concepts in sequential order: "stood", "field", "looking". \\
CommonGen, length & Generate a sentence that mentions all of these concepts in sequential order with the word count of 10, punctuation ignored: "stood", "field", "looking". \\
Gigaword, length & Given the text "japan 's nec corp. and UNK computer corp. of the united states said wednesday they had agreed to join forces in supercomputer sales .", summarize the aforementioned text in a single phrase with the word count of 6. \\
IATE / Wiktionary & Translate from English to German using terminology "Interview":\textbackslash{}n\textbackslash{}nEnglish: That is what the Hollywood star has made abundantly clear in an interview.\textbackslash{}nGerman: \\
StoryCloze, position & Given the first three sentences of the story "My friends all love to go to the club to dance. They think it's a lot of fun and always invite. I finally decided to tag along last Saturday." and two endings "My friends decided to keep inviting me out as I am so much fun." and "The next weekend, I was asked to please stay home.", infill the missing fourth sentence and choose the correct ending from the two. \\ \bottomrule
\end{tabular}
\caption{Examples of Natural Language Instructions}
\label{table:nli_example}
\end{table*}

\end{document}